# Search-based Kinodynamic Motion Planning for Omnidirectional Quadruped Robots*

Pei Wang, Xiaoyu Zhou, Qingteng Zhao, Jun Wu, Qiuguo Zhu[1]

*Abstract*—Autonomous navigation has played an increasingly significant role in quadruped robot system. However, most existing works on quadruped robots navigation using traditional search-based or sample-based methods do not consider the kinodynamic characteristics of quadruped robots, generating kinodynamically infeasible parts, that are difficult to track. In this paper, we introduce a complete navigation system considering the omnidirectional abilities of quadruped robots. First, we use kinodynamic path finding method to obtain smooth, dynamically feasible, time-optimal initial paths and add collision cost as a soft constraint to ensure safety. Then the trajectory is refined by the timed elastic band (TEB) method based on the omnidirectional model of quadruped robots. The superior performance of our work is demonstrated through simulating and real-world experiments on our quadruped robot Jueying Mini.

## I. INTRODUCTION

Legged robots have better mobility and versatility than wheeled or tracked vehicles in complex environments such as rough terrains, [1], [2], [3]. Thus, some research groups make use of the strong omnidirectional and flexible locomotive capabilities of legged robots to realize autonomous navigation. However, this is still a challenging task because of difficulties in modeling and motion planning.

Current approaches to motion planning and navigation on legged robots often focus on foothold planning [4], [5], [6], and torso-based planning [7], [8], [9], [10]. Foothold planning considers the selection of foot-placements according to dynamics model of legged robot, while torso-based planning avoids complex dynamic modeling of legged robot with a virtual body model, which separates planning and control problems, and works well with fast gait-based locomotion.

Recent work has shown the advantage of torso-based planning on real robots. In [4], quadruped StarlETH uses the RRT* algorithm to find a global path and considered the whole robot footprint, instead of the foothold. And the method carries out footprint planning many times, which is time consuming. Big Dog uses a variation of the A* algorithm for the whole body path planning and a spline algorithm for path smoothing [5]. ANYmal features the A* algorithm for pose graph planning and the RRT* algorithm for traversability planning [6]. The quadruped robot in [7] uses the Dijkstra algorithm for global planning and the equivalent virtual body model for obstacle avoidance. The torso-based planning methods mentioned above use position-only methods to plan rough global paths which are not precise or energy efficient enough, also uneasily tracked. In order to solve these problems, kinodynamic path finding method [11], [12] can be applied on quadruped robots for energy efficient, easily tracked and kinodynamic feasible prior trajectories while maintaining self-stabilization property of gait-based locomotion. In [13], a data driven approach is used to learn the kinodynamic model of the quadruped robot, and then apply it on path finding for energy efficient navigation. The method is verified in the 2d flat ground simulation only, but not in real world. Similar to [13], we use kinodynamic path finding instead of position-only methods, but we use 'two-stage' planning, the front-end and the back-end, to simplify the complex modeling problems. And we incorporate footprint planning into global planning with the use of costmap by soft constraints to avoid multiple planning. Finally, we test our method in the simulation as well as the quadruped robot platform Jueying Mini, with consideration of its omnidirectional motion ability. The main contribution of this paper can be concluded as

1) A kinodynamic path finding method is used in the front-end, instead of geometric graph searching, for energy efficient, collision-free, kinodynamic feasible, and time-optimal trajectories. These trajectories are easily tracked and can reduce the burden of back-end for trajectory optimization. Hard and soft constraints with costmap are introduced to ensure safe front-end searching.

2) We use the timed elastic band (TEB) method under omnidirectional locomotion model in the back-end for further trajectory optimization and add constraints according to actual physical parameters and different locomotive abilities (in the forward-backward and lateral directions of our quadruped robot).

The rest part of this paper is organized in the following way. Section II describes the navigation system's structure and formulates the navigation problem. Section III introduces the Kinodynamic A* algorithm and describes hard and soft constraints applied to the motion uncertainty optimization problem. In Section IV, we combine TEB and omnidirectional methods for quadruped robots to refine the front-end path planning. Section V describes the experiments that we used to verify the robot's performance and presents results. In Section VI, we summarize our work.

*This work was supported by NSFC 62088101 Autonomous Intelligent Unmanned Systems.

Qiuguo Zhu is with the State Key Laboratory of Industrial Control Technology and Institute of Cyber-Systems and Control, Zhejiang University, Zhejiang, 310027, China (corresponding author; email:qgzhu@zju.edu.cn).

Pei Wang, Xiaoyu Zhou, Qingteng Zhao, Jun Wu are with the State Key Laboratory of Industrial Control Technology and Institute of Cyber-Systems and Control, Zhejiang University, Zhejiang, 310027, China

## II. PROBLEM FORMULATION

### A. System Overview

The planning system framework is shown in Fig. 1. The framework has three important components: Costmap, Global planner, and Local planner. Costmap takes the sensor input of the real environment and inflates costs on a 2D occupancy grid map. Global planner generates an initial trajectory, containing kinodynamic information, from the current position to goal position. Local planner obtains the robot's real-time location from the state estimator, provides a controller that connects path to robot, generates a trajectory using TEB, and sends motion commands to robot control system.

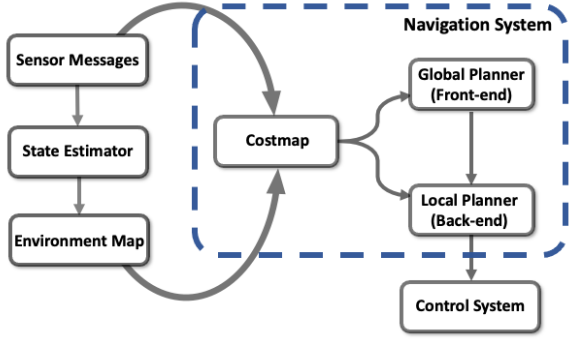

Figure 1. The structure of navigation system

### B. Problem Formulation and Assumptions

Jueying Mini [14] is a quadruped robot that has the ability to move in a wide variety of complex environments. The control system guarantees the stability, and provides omnidirectional mobility to the robot. Since motion planning is in two dimensional regions, and roll and pitch can be ignored while their values remain small, only forward-backward, lateral, and yaw motions are considered in our framework. This allows translation and rotation to be defined independently of one other and simplifies the planning problem.

## III. KINODYNAMIC A* PATH FINDING METHOD

### A. Motion Primitives Generation

Based on the assumption in Section II.B, each axis component of trajectory can be described independently. For saving computing resources to support online planning, we only focus on the $[x, y]$ in the front end. Let $x(t) \in \chi \subset \mathsf{R}^4$ be a system state, consisting of position $p(t)$ and its derivatives. Thus,

$$x(t) := \begin{bmatrix} p^\mathrm{T}(t) & \dot{p}^\mathrm{T}(t) \end{bmatrix}^\mathrm{T}$$

$$p(t) := \begin{bmatrix} p_x(t) & p_y(t) \end{bmatrix}^\mathrm{T}, p_d(t) = \sum_{k=0}^{K} a_k \frac{t^k}{k!}, d \in \{x, y\} \quad (1)$$

where $a_k$ represents polynomial coefficients. The velocity is denoted by $v(t) := \dot{p}(t)$, and also acceleration is denoted by $a(t) := \ddot{p}(t)$, and the control input $u(t) = a(t) \in U := [-u_{max}, u_{max}]^2 \subset \mathsf{R}^2$. The state space model can be described as

$$\dot{x} = Ax + Bu,$$

$$A = \begin{bmatrix} 0 & I_2 \\ 0 & 0 \end{bmatrix}, B = \begin{bmatrix} 0 \\ I_2 \end{bmatrix} \quad (2)$$

The solution for the equation is expressed as

$$x(t) = e^{At}x(0) + \int_0^t e^{A(t-\tau)}Bu(\tau)d\tau \quad (3)$$

To generate trajectories that consider more than the shortest geometric distance considered by the traditional A* algorithm, i.e., that are also smooth, collision-free, dynamically feasible, and optimal (in time and control), the qualities of the trajectory can be expressed as

$$J(\phi) = \int_0^T u^2(t)dt = \int_0^T a^2(t)dt \quad (4)$$

where $\phi$ denotes the trajectory. By taking time into consideration, the cost function is refined to

$$J(T) = J(\phi) + \rho T = \int_0^T u^2(t)dt + \rho T \quad (5)$$

where $\rho$ is the parameter which determines the relative importance of the duration versus its smoothness.

The problem defined by (5) is a linear quadratic minimum-time problem [15]. To convert the optimization problem into a graph searching problem [12], we used lattice discretization $U_M := \{u_1, u_2, ..., u_M\}$, and each control input $u_m \in \mathsf{R}^2$ is a vector in the x-y plane, which is applied for a short duration $\tau$. A discretization step $d_\mu$ was introduced to get $\mu = u_{max}/d_\mu$ samples along each axis $[0, u_{max}]$. Then, the discretized set of $(2\mu+1)^2$ primitives was $\left\{-u_{max}, -\frac{\mu-1}{\mu}u_{max}, ..., 0, ..., \frac{\mu-1}{\mu}u_{max}, u_{max}\right\}$. Since $\tau$ is short, we treat the control input as a constant vector $u_m$. Using the initial state $x_0 := \begin{bmatrix} p_0^\mathrm{T} & v_0^\mathrm{T} \end{bmatrix}^\mathrm{T}$, another form of $p_d(t)$ is written as

$$p_d(t) = u_m \frac{t^2}{2} + v_0 t + p_0 \quad (6)$$

With both duration and control input are known and fixed, we calculate the actual cost of a motion primitive as $(\|u_m\|^2 + \rho)\tau$.

Similar to traditional A*, the formulation of Kinodynamic A* algorithm has two parts, the actual cost and heuristic cost. Thus, the evaluation function $f$ is as followed

$$f = g + h \quad (7)$$

where $g$ represents the actual cost from start state to current state, and $h$ represents the heuristic cost. Thus, the actual cost $g$ from the start state to the current state is accumulated as followed.

$$g = \Sigma \left( \|u_m\|^2 + \rho \right) \tau \tag{8}$$

*B. Heuristic Function*

A suitable heuristic function reduces unnecessary expansion and results in faster searching. The distance between the current state and the goal state is heuristic for the traditional A* algorithm. Since the evaluation of $g$ has changed, and the complexity of the Kinodynamic A* algorithm is higher than that of A*, it is essential to design an admissible and tight heuristic function to speed up node expansion. By minimizing $J(T)$ from the current state to the goal state, using the Pontryagins minimum principle [16], we get

$$p_d^*(t) = \frac{1}{6}\alpha_d t^3 + \frac{1}{2}\beta_d t^2 + v_{dc} t + p_{dc}$$

$$\begin{bmatrix} \alpha_d \\ \beta_d \end{bmatrix} = \frac{1}{T^3} \begin{bmatrix} -12 & 6T \\ 6T & -2T^2 \end{bmatrix} \begin{bmatrix} p_{dg} - p_{dc} - v_{dc}T \\ v_{dg} - v_{dc} \end{bmatrix}$$

$$J^*(T) = \sum_{d \in \{x,y\}} \left( \frac{1}{3}\alpha_d^2 T^3 + \alpha_d \beta_d T^2 + \beta_d^2 T \right) \tag{9}$$

where $p_{dc}$, $v_{dc}$, $p_{dg}$, $v_{dg}$ are the position and velocity of the current state and position and velocity of the goal state, respectively. And the heuristic function $h$ can be described as

$$h = J^*(T) \tag{10}$$

The heuristic function $h$ is only related to $T$. To minimize $J^*(T)$ for the optimal $T$, we need to obtain its extremum by making $\partial J^*(T)/\partial T = 0$. Denoting the root as $T_h$, we get

$$h = J^*(T_h) \tag{11}$$

Thus, the complete form of the evaluation function $f$ is expressed as

$$f = g + h = \Sigma \left( \|u_m\|^2 + \rho \right) \tau + J^*(T_h) \tag{12}$$

*C. Collision and Dynamic Feasible Check*

We aim to find a collision-free and dynamic feasible trajectory from the start state to the goal state. Hence, it is necessary to verify collision and dynamic constraints during the search process.

The environment is described with a two-dimensional occupancy grid map. The costmap, in which each grid has a value describing the probability of collision, can be generated based on the occupancy grid map. A set of positions that the system can traverse along the trajectory can be sampled using the cost map. We define a lethal cost, related to the size of robot, that represent the collision boundary as a hard constraint. For the duration $\tau$, we need to ensure that the cost of each grid, corresponding to positions $p(t_i)$ for all $i \in \{0,...,I\}$, $t_i \in [0,\tau]$, is no more than the lethal cost. The selection of $I$ should guarantee that the maximum distance between two adjacent sampling points does not exceed costmap resolution $R$ by setting the condition $\tau v_{max} / I \geq R$.

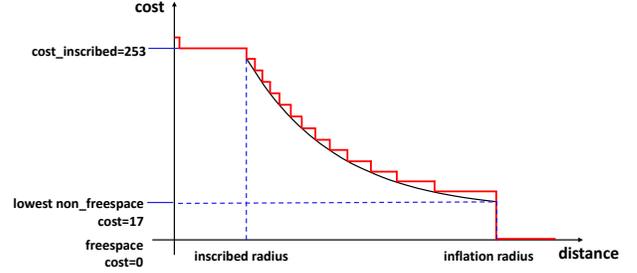

Figure 2. Trend of cost value in the costmap.

The way to determine that the primitive satisfied the dynamic constraints, is to find the maximum and minimum derivatives, such as velocity and acceleration, during $\tau$. Because the derivatives are polynomial function of time, we can easily obtain their extrema and check if they are within the constraints.

*D. Soft Constraint for Safety*

Although we obtain a collision-free trajectory with hard constraints, considering motion uncertainty, we prefer a trajectory that is as far away from obstacles as possible. Artificial Potential Field (APF) [17] is an efficient and commonly used method to maintain a path away from obstacles. However, it ignores dynamic constraints during re-optimization, and the resulting trajectory is often easily trapped in undesired local minima. Liu [18] models motion uncertainty as a soft constraint through the collision cost with an expression form of APF, but it requires a prior trajectory, from which the resulting trajectory is constrained to be within a tunnel.

A collision cost $J_c(\phi)$ was added to (5) to become $M(\phi,T)$ as a soft constraint during the searching process:

$$M(\phi,T) = J(\phi) + \rho T + \rho_c J_c(\phi) \tag{13}$$

where $\rho_c$ is the weight coefficient about collision description and $J_c(\phi)$ is defined according to the trajectory

$$J_c(\phi) = \int_\phi F(s)ds, s \in \mathsf{R}^2 \tag{14}$$

$F(s)$ is the cost in position, and it depends on the costmap value

$$F(s) = \begin{cases} 0, & l(s) \geq l_2 \\ C(l(s)), & l_2 > l(s) \geq l_1 \\ C_{max} & l(s) < l_1 \end{cases} \quad (15)$$

where $l(s)$ is the distance between the position $s$ and the nearest obstacle, and $C_{max}$ is the maximum value of $C(l)$. The inflation radius $l_2$ can be affected by the environment and safety demands. Points outside the inflation radius are considered safe. The inscribed radius of the robot is $l_1$. For the points between $l_1$ and $l_2$, the cost is defined by

$$C(l) = C_{max} e^{-\lambda_c (l-l_1)} \quad (16)$$

where $\lambda_c$ determines the decreasing rate of cost value while $l \in [l_1, l_2]$. Different from previous works [17], [18], gradient of cost function is not continuous and the range and trend can be adjusted by the inflation radius and the parameter $\lambda_c$, which may result in sharp decreases at the boundary (Fig. 2).

In practice, we can calculate the accumulated cost of each primitive by sampling. Similar to the process of collision checks, we obtain a set of $I_c$ dense points along primitives during time $\tau$. $I_c$ is determined by

$$I_c = \frac{v_{max} \tau}{R} \quad (17)$$

where $R$ is the resolution of the costmap. Start and end points of a primitive are included, so that the time step $dt := \tau / (I_c - 1)$. And the integral can be discretized as

$$\int_\phi F(s) ds \approx \sum_{i=0}^{I_c - 1} F(p_{i_c}) \|v_{i_c}\| dt \quad (18)$$

where $p_{i_c}$ and $v_{i_c}$ are the position and velocity at time $i_c \cdot dt$. Therefore, the evaluation function that produces a trajectory away from obstacles, ensuring safety of robots, is updated to

$$f = g + h + \rho_c c_{collision}$$

$$c_{collision} = \sum_{i=0}^{I_c - 1} F(p_{i_c}) \|v_{i_c}\| dt \quad (19)$$

and the influence of collision cost can be adjusted by adjusting the weight $\rho_c$.

## IV. Timed Elastic Band Trajectory Optimization

The trajectory generated by Kinodynamic A* provides not only a collision-free path but also time information, with which we can refine the prior path for a smoother and safer trajectory and take *yaw* into account using the TEB approach. TEB is based on elastic band approach, defined by $B$: a sequence of $n$ robot poses $s_i = [x_i, y_i, \theta_i]^T \in \mathbb{R}^2 \times S^1$ and $n-1$ time intervals $\Delta T_i$. And $x_i$, $y_i$ is the position, while $\theta_i$ is the orientation of the robot in global frame. These can be written as

$$Q = \{s_i\}_{i=0...n} \; n \in \mathbb{N}', \; \tau = \{\Delta T_i\}_{i=0...n-1}$$

$$B := (Q, \tau) \quad (20)$$

Because of the robot's omnidirectional properties, we obtain expressions for dynamic constraints that differ from those in [19]:

$$\Delta s_i = \begin{pmatrix} \Delta x_i \\ \Delta y_i \\ \Delta \theta_i \end{pmatrix} = \begin{pmatrix} x_{i+1} - x_i \\ y_{i+1} - y_i \\ \theta_{i+1} - \theta_i \end{pmatrix} \quad (21)$$

where $\Delta s_i$ contains the distance between consecutive positions and the angular change between the two position vectors. We convert $\Delta s_i$ in time $\Delta T_i$ from the world coordinate system to the robot coordinate system using

$$dx_i = \Delta x_i \cos\theta_i + \Delta y_i \sin\theta_i$$

$$dy_i = -\Delta x_i \sin\theta_i + \Delta y_i \cos\theta_i \quad (22)$$

after which, linear and angular velocity and acceleration can then be obtained as

$$v_{ix} = \frac{dx_i}{\Delta T_i}, v_{ix} \in [-v_{x\min}, v_{x\max}] \; v_{iy} = \frac{dy_i}{\Delta T_i}, v_{iy} \in [-v_{y\max}, v_{y\max}]$$

$$a_{ix} = \frac{v_{(i+1)x} - v_{ix}}{(\Delta T_i + \Delta T_{i+1})/2}, a_{ix} \in [-a_{x\max}, a_{x\max}]$$

$$a_{iy} = \frac{v_{(i+1)y} - v_{iy}}{(\Delta T_i + \Delta T_{i+1})/2}, a_{iy} \in [-a_{y\max}, a_{y\max}]$$

$$\omega_i = \frac{\Delta \theta_i}{\Delta T_i}, \omega_i \in [-\omega_{\max}, \omega_{\max}]$$

$$\alpha_i = \frac{\omega_{i+1} - \omega_i}{(\Delta T_i + \Delta T_{i+1})/2}, \alpha_i \in [-\alpha_{\max}, \alpha_{\max}] \quad (23)$$

A total objective function

$$R(B) = \sum_k \gamma_k R_k(B)$$

$$B^* = \arg\min_B R(B) \quad (24)$$

where $B^*$ represents the optimal TEB. And component objective functions $R_k(B)$ contain constraints about the minimum time, attraction of global path, repulsion of obstacles with respect to trajectory, and dynamic limits, such as velocity and acceleration [20]. What's more, the different locomotion capabilities of quadruped robots in forward, backward and lateral directions decide the weight $\gamma_k$ of the corresponding $R_k(B)$. And a new constraint about *yaw* is added to the total objective function to minimize the change of

*yaw* in order to save energy. Finally, TEB algorithm adopts the g2o-framework to optimize and get solutions.

## V. EXPERIMENTAL RESULTS

### A. Platform Details

Jueying Mini has four legs with 12 actuators, and each leg has 3 degrees of freedom and each joint has an expanded range of motion. (Fig. 3) The kinodynamic constraints in navigation of Jueying Mini are shown in Table I. Jueying Mini is equipped with Velodyne VLP-16 LiDAR and IMU for sensing and state estimation. All software modules, including state estimation, mapping, and planning, run on a four-core 2.80 GHz processor with a 256 GB hard disk and a 8 GB RAM.

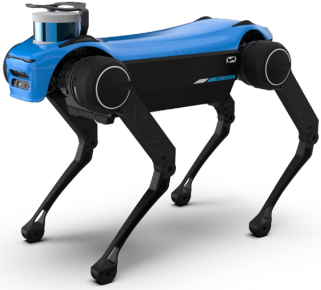

Figure 3. The Jueying Mini quadruped robot.

TABLE I. KINODYNAMIC CONSTRAINTS IN EXPERIMENTS

| Name | Value | Unit |
|---|---|---|
| $v_{x\ max}$ | 0.75 | |
| $v_{x\ min}$ | 0.10 | m/s |
| $v_{y\ max}$ | 0.20 | |
| $\omega_{max}$ | 0.70 | rad/s |
| $a_{x\ max}$ | 1.00 | |
| $a_{y\ max}$ | 0.17 | m/s$^2$ |
| $\alpha_{max}$ | 0.52 | rad/s$^2$ |

### B. Simulations

*1) Search-Based Planning Performance:* To evaluate the Kinodynamic A* algorithm, we compare its performance with that of the traditional A* path planning algorithm. The results cannot be obtained directly, since the path generated by traditional A* does not contain any kinodynamic information. Instead, we test the generated trajectories in real execution. With the only difference between experiments lying in the front-end, we can verify the performance between Kinodynamic A* and traditional A*.

The experiment was performed on a simulated irregular polygon map approximately 30 m × 30 m, with data recorded during navigation to evaluate the performance of two methods. As shown in Table II, the start and goal positions of four groups were randomly selected. For Kinodynamic A*, the effort of trajectory and the total running time for each experiment was better than those of the traditional A*. Kinodynamic A* resulted in a smoother and more easily tracked prior path, while the path of traditional A* had many dynamically infeasible parts, which required more energy and time for the local planner to optimize and adjust. Therefore, we demonstrate that Kinodynamic A* algorithm can finally lead to a smooth and fast trajectory to execute.

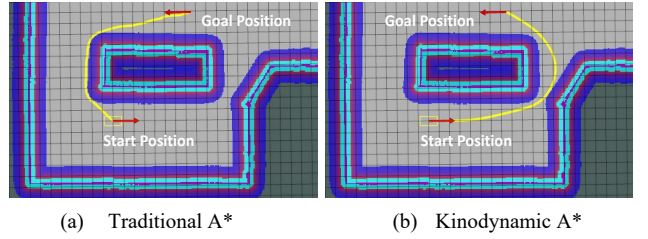

Figure 4. Dynamic feasibility performance. The yellow rectangle represents the outline, the yellow line represents the path, and the red arrow represents the position vector of the robot.

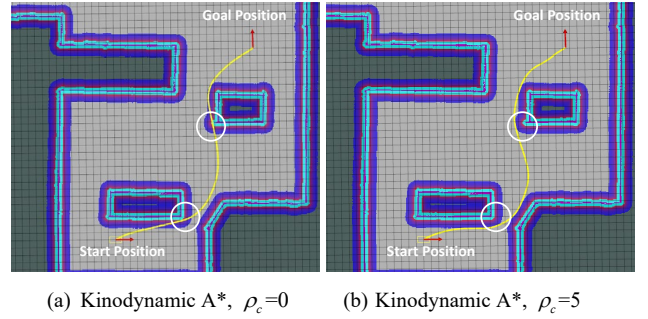

Figure 5. Performace after adding a collision cost. The collision cost can be achieved by adjusting $\rho_c$. The yellow rectangle represents the outline, the yellow line represents the path, and the red arrow represents the position vector of the robot.

TABLE II. COMPARISON OF PATH PLANNING ALGORITHMS

| | Total Running Time (s) | | Effort J | |
|---|---|---|---|---|
| | *Kinodynamic A\** | *Traditional A\** | *Kinodynamic A\** | *Traditional A\** |
| 1 | **27.362** | 31.794 | **4.7630** | 5.5047 |
| 2 | **16.298** | 25.395 | **5.1730** | 6.8421 |
| 3 | **25.338** | 31.000 | **5.1201** | 5.8651 |
| 4 | **25.975** | 25.594 | **4.1946** | 7.4720 |

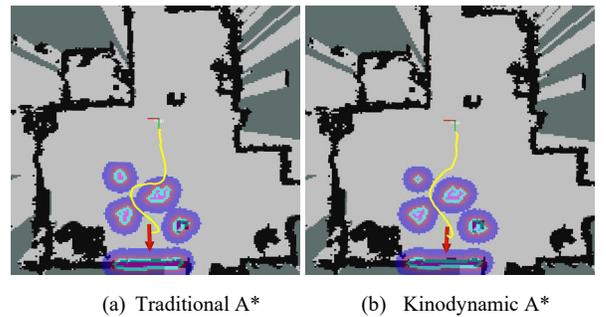

Figure 6. The actual trajectories of the robot. The yellow line represents the actual trajectory of the robot and the red arrow represents the position vector of the robot.

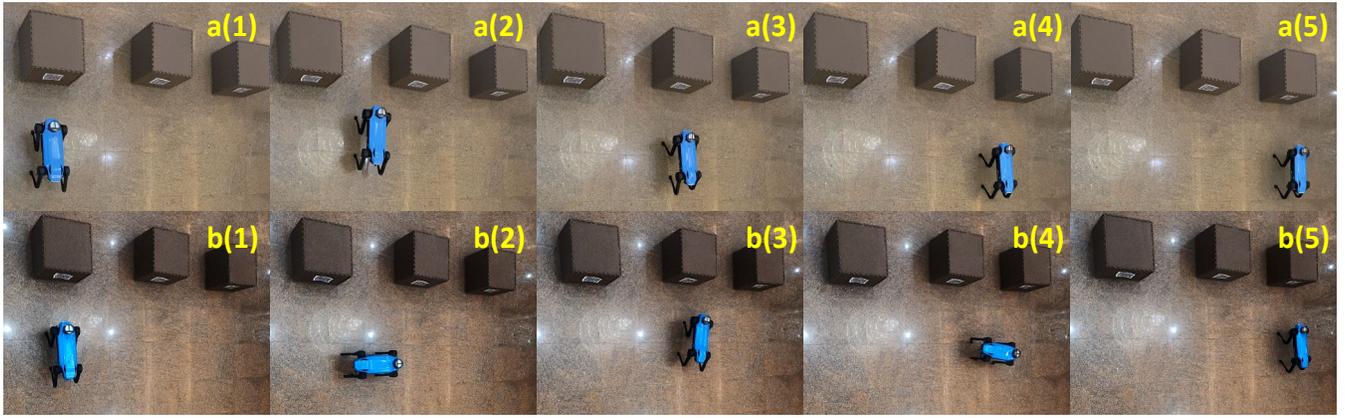

Figure 7. Three parallel points are set up to simulate a situation where the robot is performing actual tasks. The robot needs to face toward the object and stay for 2s. The upper pictures shows the performance of the omnidirectional method, and the lower one shows the performance of the non-omnidirectional method.

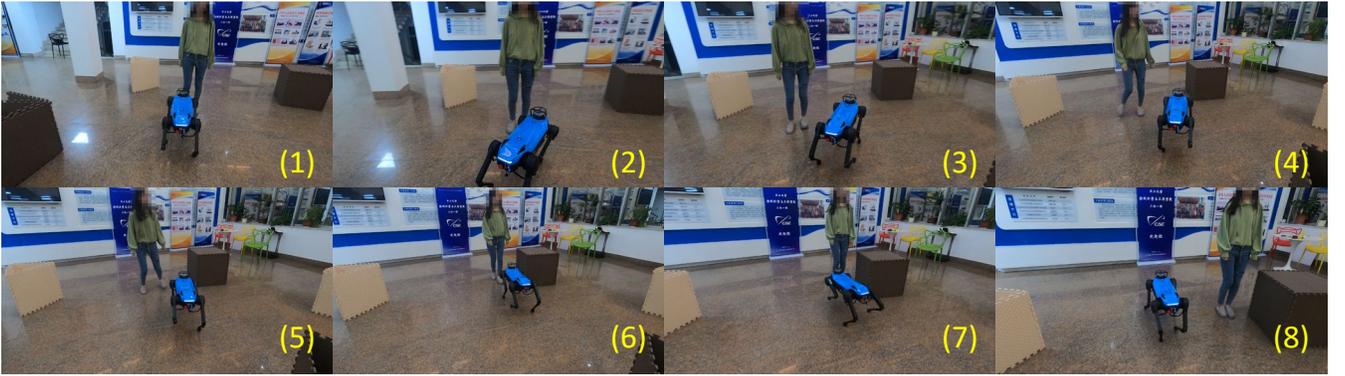

Figure 8. Dynamic obstacle avoidance performance.

Another advantage of Kinodynamic A* is that it takes kinodynamic characteristics into consideration. As shown in Fig.4, the global planner using A* contains no motion information, it ignores the initial state, especially in re-planning. This may result in trajectories that are unsuitable. However, Kinodynamic A* considers the robot's initial state and avoids the unnecessary cost of turning around.

*2) Collision Cost Performance:* We verified the contribution of soft collision constraints. Fig. 5(a) shows results of the method without soft constraints, in which the robot may hit the obstacle inside the circle region. Fig. 5(b) shows the results of the method with collision costs, in which the robot avoids the obstacles by walking a safer path.

*C. Experiments*

*1) Search-Based Planning Performance:* We conducted the experiment in an incompletely known environment with unexpected obstacles, unknown in the initial costmap. Fig. 6 shows the actual trajectories of the robot. The Kinodynamic A* algorithm provided an easily tracked prior trajectory which led to a better final trajectory, while the trajectory produced by the traditional A* algorithm had many dynamically infeasible parts that made tracking difficult for the local planner.

*2) Omnidirectional Method Performance:* We compared the performance of TEB under omnidirectional model assumption the non-omnidirectional model assumption in an indoor environment as shown in Fig. 7. The overall running time of each experiment was compared. Because of the difference in forward and lateral locomotion abilities, we set a greater weight for the forward direction. As shown in Table III and Fig. 7, the omnidirectional method was flexible and fast. Because the omnidirectional motion characteristics of the quadruped robot were considered, our method enabled robot to transfer flexibly between the parallel points by side shifting.

TABLE III. COMPARISON OF OMNIDIRECTIONAL AND NONOMINIDIRECTIONAL METHODS

|  | Time(s) | |
| --- | --- | --- |
|  | *Omnidirectional method* | *Non-omnidirectional method* |
| To first via point | 11 | 11 |
| To second via point | **10** | 20 |
| To third via point | **11** | 23 |
| Total time | **32** | 54 |

*3) Dynamic Obstacle Avoidance:* We test the robot using Kinodynamic A* and TEB with the omnidirectional model in a dynamic and completely unknown environment as shown is Fig.8. Perception range limits prevented some obstacles from being considered in the initial global plan, which was a challenge to planning, requiring continuous and rapid re-planning to avoid sudden dangers. As shown in Fig. 8, our method can generate feasible trajectories and allow the robot to flexibly react. Additional details can be seen in our accompanying video.

## VI. Conclusion

In this paper, we describe the design of a complete autonomous navigation system for quadruped robots that takes into account their ability for flexible, powerful, and stable omnidirectional locomotion. We use a framework with a global planner and a local planner, to find a path from start to goal and calculate velocity commands for robot to execute. The global planner uses Kinodynamic A* to find a smooth, safe, kinodynamically feasible, and minimum-time prior path, which contains hard and soft constraints to guarantee the clearance and dynamic feasibility. The initial path is refined by local planner using TEB method with an omnidirectional model. Finally, we verify the effectiveness and feasibility of our method through experiments and prove that our quadruped robot can perform more flexible movements and react rapidly to changes in complex environments.